\documentclass[a4paper]{article}
\usepackage{idsiatechrep}      
\usepackage{times}

\usepackage{amssymb,amsmath}
\usepackage[ruled]{algorithm2e}
\usepackage{subfig}
\usepackage{graphicx}

\newcommand{\Cc}{{\mathcal C}}           
\newcommand{\Rr}{{\mathcal R}}  
\newcommand{\Aa}{{\mathcal A}} 
\newcommand{\Ss}{{\mathcal S}}
\newcommand{\Pp}{{\mathcal P}}     

\newcommand{\argmax}[1]{\arg\!\max_{\!\!\!\!\!\!#1}}

\title{T-Learning}
\author{Vincent Graziano, Faustino Gomez, Mark Ring \\ 
 ~and~J\"{u}rgen Schmidhuber}
\date{December 2011,}

\reportnumber{XX-11}

\reportmonth{December}
\reportyear{2011}

\begin{document}

\makecover         
\maketitle

\begin{abstract}
Traditional Reinforcement Learning (RL) has focused on problems
involving many states and few actions, such as simple grid
worlds. Most real world problems, however, are of the opposite type,
Involving Few relevant states and many actions. For example, to return
home from a conference, humans identify only few subgoal states such as lobby,
taxi, airport etc. Each valid behavior connecting two such states can
be viewed as an action, and there are trillions of them. Assuming the
subgoal identification problem is already solved, the quality of any
RL method---in real-world settings---depends less on how well it
scales with the number of states than on how well it scales with the
number of actions. This is where our new method T-Learning excels, by
evaluating the relatively few possible transits from one state to
another in a policy-independent way, rather than a huge number of
state-action pairs, or states in traditional policy-dependent
ways. Illustrative experiments demonstrate that performance improvements of T-Learning over Q-learning can be arbitrarily large.

\end{abstract}

\section{Motivation and overview} 

Traditional Reinforcement Learning (RL) has focused on problems
involving many states and few actions, such as simple grid
worlds. Most real world problems, however, are of the opposite type,
involving few relevant states and many actions. For example, to return
home from a conference, humans identify only few subgoal states such as lobby, taxi, airport etc. Each valid behavior connecting two such states can be viewed as an action, and there are trillions of them. Assuming the subgoal identification problem is already solved by a method outside the scope of this paper, the quality of any RL method---in real-world settings---depends less on how well it scales with the number of states than on how well it scales with the number of actions.

Likewise, when we humans reach an unfamiliar state, we generally resist testing every
possible action before determining the good states to transition to.
We can, for example, observe the state transitions that other humans progress
through while accomplishing the same task, or reach some rewarding state by happenstance.
Then we can focus on reproducing that sequence of states. That is, we are able to first identify a task before acquiring the skills to reliably perform it.
Take for example the task of walking along a balance beam. 
In order to traverse the length of the beam without falling, a precise action
must be chosen at every step from a very large set of possibilities.
The probability of failure is high because almost all actions at every step
lead to imbalance and falling, and therefore a good deal of training is
required to learn the precise movements that {\em reliably} take one across.
However, throughout the procedure the desired trajectory of states is well
understood; the more difficult part is achieving them reliably.

Reinforcement-learning methods that learn action values, such as
$Q$-learning, Sarsa, and TD(0) are guaranteed to converge to the
optimal value function provided all state-action pairs in the underlying MDP are visited infinitely often. These methods therefore can converge extremely slowly in environments with large action spaces.

This paper introduces an elegant new algorithm that automatically focuses
search in action space by learning state-transition values independent of
action.
We call the method $T$-learning, and it represents a novel off-policy approach
to reinforcement learning.
$T$-learning is a temporal-difference (TD) method~\cite{rl:98}, and
as such it has much in common with other TD-methods, especially action-value
methods, such as Sarsa and $Q$-learning~\cite{watkins,qlearn:92}.
But it is quite different.
Instead of learning the values of state-action pairs as action-value methods
do, it learns the values of state-state pairs (here referred to as
\emph{transitions}).

The value of the transitions between states is recorded explicitly, rather
than the value of the states themselves or the value of state-action pairs.
The learning task is decomposed into two separate and independent components:
(1) the learning of transition values, (2) the learning of the optimal
actions.
The transition-value function allows high payoff transitions to be easily
identified, allowing for a focused search in action space to discover those
actions that make the valuable transitions reliably.

Agents that learn the values of state-transitions can exhibit markedly
different behavior from those that learn state-action pairs.
Action-value methods are particularly suited to tasks with small action spaces
where learning about all state-action pairs is not much more cumbersome than
learning about the states alone.
However, as the size of the action space increases, such methods become less
feasible.
Furthermore, action-value methods have no explicit mechanism for identifying
valuable state transitions and focusing learning there.
They lack an important---real-world---bias: that valuable state transitions can often be
achieved with high reliability.
As a result, in these common situations, action-value methods require
extensive and undue search before converging to an optimal policy.
$T$-learning, on the other hand, has an initial bias: it presumes the
existence of reliable actions that will achieve any valuable transition yet
observed.
This bias enables the valuable transitions to be easily identified and search
to be focused there.
As a result, the difficulties induced by large action spaces are significantly
reduced.

\section{Environments requiring precision}\label{sec:skillenv}

\newcommand{\CalS}{\mathcal{S}}
\newcommand{\CalP}{\mathcal{P}}
\newcommand{\CalR}{\mathcal{R}}

Consider the transition graph of an MDP, where the vertices of the graph are
the states of the environment and the edges represent transitions between
states.
Define a function $\tau \colon \Ss \to \Ss$ that maps $s$ to the neighboring
vertex $s'$ whose value under the optimal policy, $V^*(s')$ is the highest of
all the neighbors of $s$, where $V^*(s)$ is calculated using a given value for
$\gamma$ as though the agent had actions available in every state that can move it
deterministically along the graph of the environment.

The class of MDPs for which T-Learning is particularly suited can be described
formally as follows: If $s \overset{\tau}\mapsto s'$, then

\begin{enumerate}
\item  $E[Pr(s' | s, a\in A) > \epsilon]$, and
\item $Pr(s' | s, a^*) > 1-\epsilon,$ for some $a^* \in A$,
\end{enumerate} 

\noindent where $\epsilon$ is a small positive value.

These environments are those where specific skills can accomplish tasks
reliably.
Walking across a balance beam, for example, requires specific skills.
The first constraint ensures that the rewarding transitions are likely to be
observed.
The second constraint ensures that the transitions associated with large
reward signals can be achieved by finding a specific skill, i.e., a reliable
action.
Without this guarantee, one might never attempt to acquire certain skills
because the \emph{average} outcome during learning may be undesirable.

\begin{figure*}[th]
	\centering
        \hspace{.25cm}
       \subfloat[Basic example]{\label{fig:smallMDP}\includegraphics[width=.35\textwidth]{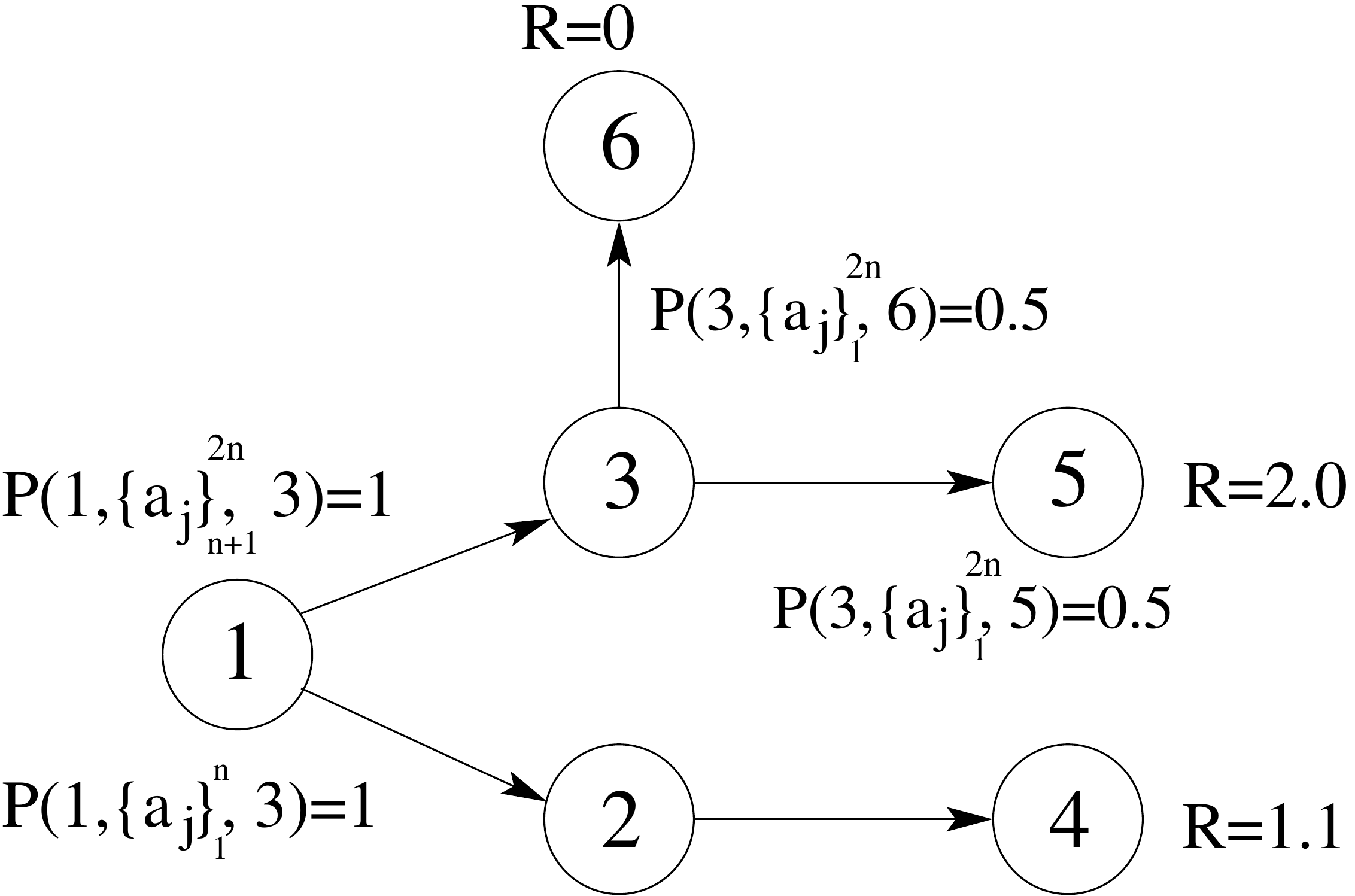}}
        \hspace{.125cm}
        \subfloat[Balance beam environment]{\label{fig:bbenv}\includegraphics[width=.55\textwidth]{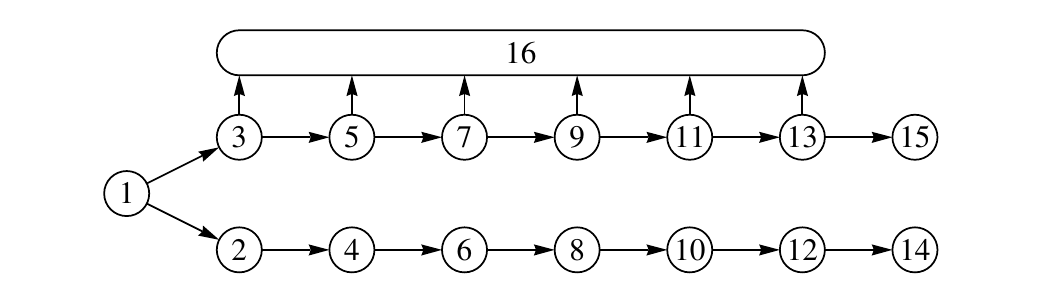}}
	\caption{Skill-based environments. Left: The agent either
      transitions to state $2$ to collect a reward of $1.1$ at state $4$
      or to state $3$ where precise skills are required. 
      The rewarding transition $3 \to 5$ can be made
      {\em reliably} only by action $a^{*}$. The other actions take the
      agent to either state $5$ or $6$ with equal likelihood. State $5$ has a reward of $2$ and state $6$ a
      reward of $0$. Right: An extension of the environment on the left.
      Both environments satisfy the {\em precision property} (see Section~\ref{sec:precision}) trivially, since there are behaviors which deterministically reach the most rewarding state. 
    }
\label{fig:skillenv}
\end{figure*}

Consider the example of Figure \ref{fig:smallMDP}.
This MDP has two parts, one requiring high skill (which yields large reward)
and one requiring low skill (which yields small reward).
Episodes begin in state $1$ and end in states $4,5$, and $6$.
There are $2n+1$ actions and the transition table is defined as follows: from
state $1$, $n$ actions, $\{a_1, \ldots, a_n\}$, take the agent to state $2$
deterministically; $n$ actions $\{a_{n+1}, \ldots, a_{2n}\}$ take the agent to
state $3$ deterministically, and one action, $a^{*} \equiv a_{2n+1}$, takes
the agent to either state $2$ or $3$ with equal probability.
All actions from state $2$ take the agent to state $4$, ending the episode.
From state $3$, $2n$ actions move the agent to either state $5$ or $6$ with equal probability,
 while action $a^{*}$ moves the
agent to state $5$ deterministically.
The agent receives a reward of $1.1$ for arriving in state $4$, and rewards of
$2$ and $0$ for arriving in states $5$ and $6$ respectively.

This example meets the criteria given above.
The rewarding transition, $3 \to 5$, is likely to be observed even before
action $a^{*}$ is discovered.
Temporal difference (TD) methods will find the optimal policy when every
state-action pair is visited infinitely often.
$Q$-learning, for example, will eventually, through exploration, discover
action $a^{*}$ at state $3$.
However, before the optimal policy is found, and after only a few episodes,
the agent will select actions that take it from state $1$ to state $2$.
This agent has no bias towards discovering the action $a^{*}$ (at state $3$),
which represents the skill required to move reliably to the rewarding state.

Agents that assign values to state-action pairs and then determine their
policies from these values cannot explicitly search for
%
%
an action that reliably makes a particular transition; rather, the rewarding
state-action pair has to be discovered as a unit.

The next section describes $T$-learning in detail.
This algorithm biases the behavior of the agent towards finding the actions that
make the most valuable transitions at each state.

\section{State transition functions}\label{sec:met}

The general reward for an MDP is a function of three variables, \[ \Rr \colon
\Ss \times \Aa \times \Ss \to \mathbb{R}. \] Most environments considered in
practice, however, take reward functions that depend on a single variable,
usually only the state of the agent. For reasons discussed below, we consider rewards as functions of state transitions,
 independent of the action taken; i.e.,
\[ \Rr \colon \Ss \times \Ss \to \mathbb{R}. \] \noindent We denote the restricted function by
$\Rr_{ss'}$, and the more general reward function by $\Rr^{a}_{ss'}$.

Recall TD(0) which learns a function $V \colon
\Ss \to \mathbb{R}$ using the following update rule,

\[ V(s) \leftarrow V(s) + \alpha [ r + \gamma V(s') - V(s) ]. \]

For a fixed policy $\pi$ this function converges to
$V^{\pi}$ which is given recursively by

\[V^{\pi}(s) = \sum_{a} \pi(s,a) \sum_{s'} \Pp^{a}_{ss'} [ \Rr^{a}_{ss'}+\gamma V^{\pi}(s')
]. \]

The next two sections present two separate learning rules.
Both learn functions that assign values to state transitions, $T \colon \Ss
\times \Ss \to \mathbb{R}$.
The first is on-policy and is essentially equivalent to TD(0).
The second is entirely {\em off-policy} and has similarity to $Q$-learning.

\subsection{An on-policy learning rule}

In the remainder of the paper, the term {\em transition functions} refers to
functions $T \colon \Ss \times \Ss \to \mathbb{R}$.
Their values are called {\em transition values} or {\em T-values}.
Consider the following update rule,
 
\[ T(s,s') \leftarrow T(s,s') + \alpha \left[ r + \gamma T(s',s'') - T(s,s')\right]. \] 

The value $T^{\pi}(s,s')$ represents the reward for the transition from $s$ to
$s'$ plus the cumulative future expected discounted reward under the given
policy.
Convergence is therefore implied by theorem that establishes the convergence of
the state value function learned by TD(0). For a fixed policy $\pi$ this
function converges to $T^{\pi}$ which is given recursively by \[T^{\pi}(s,s')
= \Rr_{ss'} + \gamma \sum_{a'} \pi(s',a') \sum_{s''} \Pp^{a'}_{s's''}
T^{\pi}(s',s''). \]

The recursive relation can be given also when
the reward is a function of three variables.\footnote{The term $\Rr_{ss'}$ would be replaced by the
expected reward for making the transition from $s$ to $s'$ under the
given policy. To find this value one would have to find the likelihood
of each action $a$ given the transition $s \to s'$. This value would
depend on $\Pp^{a}_{ss'}$, and $\pi$.} Using the restricted reward
function we have the following relations between $V^{\pi}$ and
$T^{\pi}$:

\[ T^{\pi}(s,s') = \Rr_{ss'} + \gamma V^{\pi}(s'), \] 
and
\[ V^{\pi}(s) = \sum_{a} \pi(s,a) \sum_{s'} \Pp^{a}_{ss'} T^{\pi}(s,s'). \]

One can use, for example, a one-step lookahead to select actions based on
$T(s,s')$. For example, a deterministic
policy could be given by:

\[\pi(s) = \argmax{a} \sum_{s'} \widehat{\Pp}^{a}_{ss'} T(s,s'), \]  

\noindent where $\widehat{\Pp}$ is a learned model of the
transition probabilities $\Pp$. This is similar to
  determining a policy from state values:

\[ \pi(s) = \argmax{a} \sum_{s'} \widehat{\Pp}^{a}_{ss'}
(\widehat{\Rr}^{a}_{ss'} + \gamma V(s') ), \] \noindent where it is necessary to
have a model $\widehat{\Rr}^{a}_{ss'}$ of the reward function as well.

This learning rule is on-policy. The values, as we have shown, are related to those learned by TD(0). The formulation of the rule itself is similar to the learning rule
used in SARSA. Next we introduce $T$-learning, a TD prediction method that is off-policy and analogous to $Q$-learning.

\subsection{T-Learning}

Now consider a function $T \colon \Ss \times \Ss \to \mathbb{R}$ which is learned as follows:

\[
T(s,s') \leftarrow T(s,s') + \alpha \left[ r+ \gamma \max_{s''} T(s',s'') - T(s,s') \right]. 
\]

We call this learning rule {\bf T-Learning}. This rule captures the
values associated with the best transition available. 
When the agent's behavior is determined by these values it becomes possible to  
search the action space---at the valuable states---to discover the reliable actions. Moreover,
this can be done in a straightforward and natural way.
Taking the maximum over the possible state transitions is
reminiscent of $Q$-learning; rather than capture the ideal action
associated with each state $T$-learning caches the topology of the ideal
transitions. The ideal transitions between states can be determined without having to use a model.
At state $s$ the ideal transition is simply $\arg\!\max_{\tilde{s}} T(s,\tilde{s})$. 

In the example given in Figure \ref{fig:smallMDP}, the first time the
transition from $3 \to 5$ is made, {\em regardless of the action
  selected}, the agent learns the value associated to the
transition. A subsequent backup for the transition $1 \to 3$ will make the value of
$T(1,3)$ greater than the value $T(1,2)$. The agent's policy
then shifts, preferring state $3$ to state $2$. All this can happen
before the agent discovers $a^{*}$.

\subsubsection{Appropriate environments for $T$-learning}{\label{sec:precision}

The environment needs to satisfy some niceness properties for $T$-learning to be successful. Sufficient conditions were given in
Section \ref{sec:skillenv} and occur in many real world environments. These restrictions can, however, be relaxed. We denote the function that $T$-learning converges to by $T^{\sharp}$, and the $Q$-values of the optimal policy $\pi^{*}$ by $Q^{*}$. For each state $s$ the MDP needs to satisfy the following property:

\[ \arg\!\max_{a} \sum_{s'} \Pp^{a}_{ss'} T^{\sharp}(s,s') \equiv \arg\!\max_{a} Q^{*}(s,a). \]
 
We christen this criterion the {\em precision} property. We call this the precision property because it guarantees that the valuable transitions can be made with as high reliability as needed. These reliable actions may be rare, among all possible actions, and may be considered {\em skilled} actions or behaviors. Said differently, these are MDPs where there are actions available that make the paths on the state-transition graph with the highest value (as described in Section \ref{sec:skillenv}) worth attempting.     

As an example of how the learning rule can fail, consider the
environment introduced in Figure~\ref{fig:smallMDP} without action $a^{*}$. $T$-learning will
still prefer transition $1 \to 3$ since the rule is biased towards the
payoff associated to transition $3 \to 5$. This happens because the value is
calculated independently of the specific actions available. The learning rule tacitly assumes that transitions in the transition graph can be made with arbitrarily high reliability. With $a^*$ removed from its repertoire, this assumption does not lead to the optimal policy. For this reason we restrict our discussion to environments satisfying the precision property. 

It is important to realize that action $a^{*}$ need not deterministically make the transition
$3 \to 5$. Given the same reward function action
$a^{*}$ need only make the transition $3 \to 5$ with probability greater than
$0.55$ (making the average reward when going to state $3$ greater than the $1.1$ received for transitioning to state $4$) to ensure that the converged $T$-values can be used to calculate the optimal policy.

\section{Experiments}

We compare $T$-learning to $Q$-learning in a model of a balance beam
environment (\hspace{1pt}TD(0) and other methods are discussed in Section~\ref{sec:conclusions}). The MDP has
$16$ states, the transition graph is given in Figure~\ref{fig:bbenv}. The transitions are similar to the smaller
version of this environment. We vary the number of actions, $2n+1$, throughout the experiments. The first $n$-actions move the agent
deterministically from state $1$ to $2$ and the second $n$-actions move the
agent to state $3$. Action $a^{*} = 2n+1$ transitions the agent to
either state $2$ or $3$ with equal probability. From state $2$ all
actions advance the agent deterministically along the chain, $2 \to 4 \to
6 \to 8 \to 10 \to 12 \to 14$, where the agent receives a reward of
$+1$ for reaching state $14$. The odd states represent the balance
beam. The transitions $3 \to 5 \to 7 \to 9 \to 11 \to 13 \to 15$ can
be made deterministically by action $a^{*}$. The other actions, with
equal probability, either advance the agent along the balance beam,
moving to the next odd state, or cause the agent to fall off the beam,
moving to state $16$. The agent receives no reward for reaching state
$16$ and a reward of $2$ for reaching state $15$. Episodes end at
states $14,15,16$ and begin at state $1$.

\SetFuncSty{sc}
\begin{algorithm}[t]
\DontPrintSemicolon
\caption{Action selection using $T$-values{($\epsilon,\kappa$)}}
\label{alg:policy}
 Get previous data point $\{s_{-1},a_{-1},s_{0},r_{0}\}$.\;
 Increment counters $\Cc_{sas'}$ and $\Cc_{sa}$.\;
 Current state $s = s_0$\;
	\eIf{$\mathrm{RandomReal[0,1]} < \epsilon$}{
	$\Aa_{\circ} \leftarrow \Aa$\;
	}{
	Update $\widehat{\Pp}_{s}$ as follows:\;
	\eIf{$\Cc_{sa}=0$}{
		Get number $n$ of states $s'$ transitioned to from $s$.\;
		\( p(s'|s,a) \leftarrow \begin{cases} 
                                       \kappa & \text{if } T(s,s') = \max_{s^*}T(s,s^*), \\ 
                                       (1- \kappa) / n & \text{if } s \to s' \text{ previously seen}, \\
                                       0  & \text{otherwise}. 
		\end{cases} \) \;
			}{
		$p(s'|s,a) \leftarrow \Cc_{sas'} / \Cc_{sa}$\;
		}
$\Aa_{\circ} \leftarrow \{a \in \Aa \colon \sum_{s'}\widehat{P}^{a}_{ss'}T(s,s') = \arg\max_{\tilde{a}} \sum_{s'} \widehat{P}^{\tilde{a}}_{ss'}T(s,s') \}$ \; 
	}
Select an action $a \in \Aa_{\circ}$\;
\end{algorithm}

$T$-learning is a form of TD prediction and as such it requires a separate module to generate a policy. We chose a simple model-based approach for clarity of exposition. We are interested in what $T$-learning is learning compared to what $Q$-learning is learning; ideal control (e.g., model versus model-free: actor/critic methods) does not fall into the scope of this paper.  The policy for the $T$-learning is derived from a model of the
transition matrix $\Pp^{a}_{ss'}$ and the $T$-values. The algorithm uses a one-step lookahead where new actions are selected in favor of those which fail to make the rewarding
transitions reliably. Let $\Cc_{sa}$ denote the number of times a state-action pair $(s,a)$ has been observed and $\Cc_{sas'}$ denote the number of times a transition was seen with a specific action, $(s,a,s')$. From this it is easy to generate a basic estimate $\widehat{\Pp}^{a}_{ss'} \equiv \Cc_{sas'} / \Cc_{sa}$ of the transition matrix $\Pp$. Actions are selected, in state $s$, from among those whose values $\sum_{s'} \widehat{\Pp}^{a}_{ss'} T(s,s')$ are equal the maximum.  
Actions which have yet to be taken in state $s$ are biased towards state $s^{*} = \arg\!\max_{\tilde{s}}T(s,\tilde{s})$ with a transition probability of $\kappa$. See Algorithm \ref{alg:policy} for details. For the experiment we set the
parameter 
$\kappa = 0.75$. The other parameters for both $T$-learning and
$Q$-learning are as follows: learning rate is $\alpha=0.5$, discount
factor is $\gamma = 0.85$, and exploration rate is $\epsilon=0.1$. We ran 50 trials for each of the experiments. Each trial lasted until
the policy converged.

\section{Results and Discussion}\label{sec:conclusions}

For $n=50$---a total of $101$ actions---$T$-learning required $23,511$ steps (actions executed) on average for the policy to
converge with a standard deviation of $6,580$, whereas $Q$-learning
required an average of $675,341$ steps with a standard deviation of
$284,891$, a speedup of 25 times in this relatively small
environment. 

\begin{figure}[t]
 \begin{center}
  \includegraphics[width=.4\textwidth]{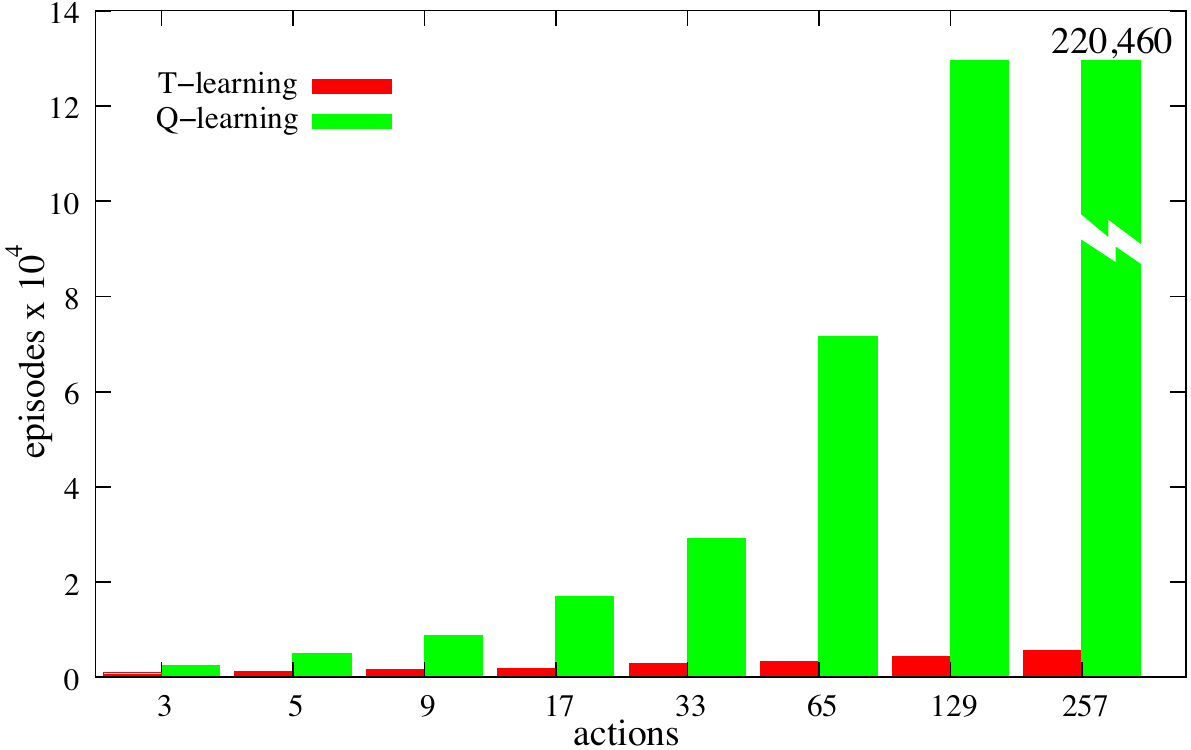}
 \end{center}
 \caption{Both methods, T-Learning and Q-Learning, see a linear increase in convergence times as the number of actions is increased (linearly).
 The ratio of episodes to convergence (Q-to-T) between the two algorithms increases approximately at a rate of $2^{.53} \approx 1.44$ with each doubling in size of the action space. T-Learning gains in performance over Q-Learning as the number of actions increase.}
  \label{fig:tvq}
\end{figure}

In Figure~\ref{fig:tvq} we see how the number of {\em episodes} to
convergence relates to the size of the action space; T-Learning
yields arbitrary speed-up factors over Q-Learning as the action space grows.

Figure~\ref{fig:behave} illustrates the key differences in behavior as a result of 
what each algorithm is learning. In the early learning stages the two algorithms exhibit the same behavior. At this point they are equally likely to traverse the beam. After a number of chance successes the $T$-learning algorithm propagates the state-transition values to state $1$. At this time it's behavior departs from the behavior exhibited by the Q-Learning algorithm-- the agent begins to favor state $3$ to state $2$. Unlike T-Learning, the Q-Learning algorithm cannot independently {\em identify the task} and {\em acquire the skill} to succeed in the task. See Figure~\ref{fig:TvQbehave}. Rather, and this represents the fundamental weakness of $Q$-learning in this environment, it's behavior is such that it always prefers the transition to state $2$ from state $1$ until it has found action $a^{*}$ in each of the odd-numbered states. Moreover, this action needs to be discovered in each of the odd states, $s$, after the value of the next odd state, $V(s') = \max_{a} Q(s',a)$, is positive. Otherwise, after the learning step, the value $Q(s,a^{*})$ will remain non-positive and thus valued as a suboptimal action. The sampling along the beam achieved by the exploration factor for Q-Learning is significantly less than than the sampling rate that T-Learning enjoys by a change in policy.

There is a horizontal asymptote for the number of episodes required for the convergence of the T-values with respect to the number of actions. See Figure~\ref{fig:tbehave}. The behavior of the agent will shift---preferring state $3$ to state $2$---around $3500$ episodes regardless of the number of actions. This is a remarkable feature of the T-Learning algorithm and might play an important role in the so-called options~\cite{Stolle02learningoptions} framework, discussed below.

Algorithms learn based on the learning rule they are wrapped around. For example, Dyna-Q~\cite{dyna} which learn a model and take advantage of planning to speed learning, still needs to first find the actions that represent the skilled movement(s) before the learning is sped-up. This is because the algorithm inherits the disadvantages of Q-Learning discussed above. However, once a well-informed model (having tried the actions representing skilled actions) is learned  the values will quickly produce the optimal policy. Also, the bottle neck of having to first discover the skilled action at state $13$ before valuing the skilled actions at other odd states will be removed. That said, planning methods can be used with T-Learning as well, and would allow for decrease in the time needed for convergence of the T-values, the agent would only have to make it across the beam a single time before shifting its policy. Using TD(0) under the hood of Dyna or Prioritized Sweeping~\cite{psweep} does not address the fundamental problem either. TD(0) is an on-policy method, and as such will learn values based on the distribution of its samples. TD(0) has results similar to Q-Learning in the balance beam environment. Models and planning methods will speed Q-Learning, but they do not address the fundamental problem of learning in MDPs with huge action spaces.

An {\em optimistic initialization of $Q$-values} does not address the heart of the matter either. For tiny actions spaces ($2$, $4$ actions) optimistic initialization puts Q-Learning roughly on par with T-Learning. In moderate sized spaces ($8$, $16$ actions) optimistic initialization took convergence time to $50\%$ of the original. With larger action spaces the number of episodes to convergence was effected less by optimistic initialization, taking about $90\%$ the original time. In general, when there are thousands upon thousands of actions it is a bad idea to have to try them all in each state. Similarly, using optimistic state-values with TD(0) with either a one-step look-ahead or an actor-critic method to generate the policy also fails to make learning significantly faster. The value of state $3$ will decrease faster than the value of state $2$, immediately nullifying the optimistic initialization; the results would be similar to those reported herein.

\begin{figure*}
  \centering
  \subfloat[Behavior during learning]{\label{fig:TvQbehave}\includegraphics[width=0.45\textwidth]{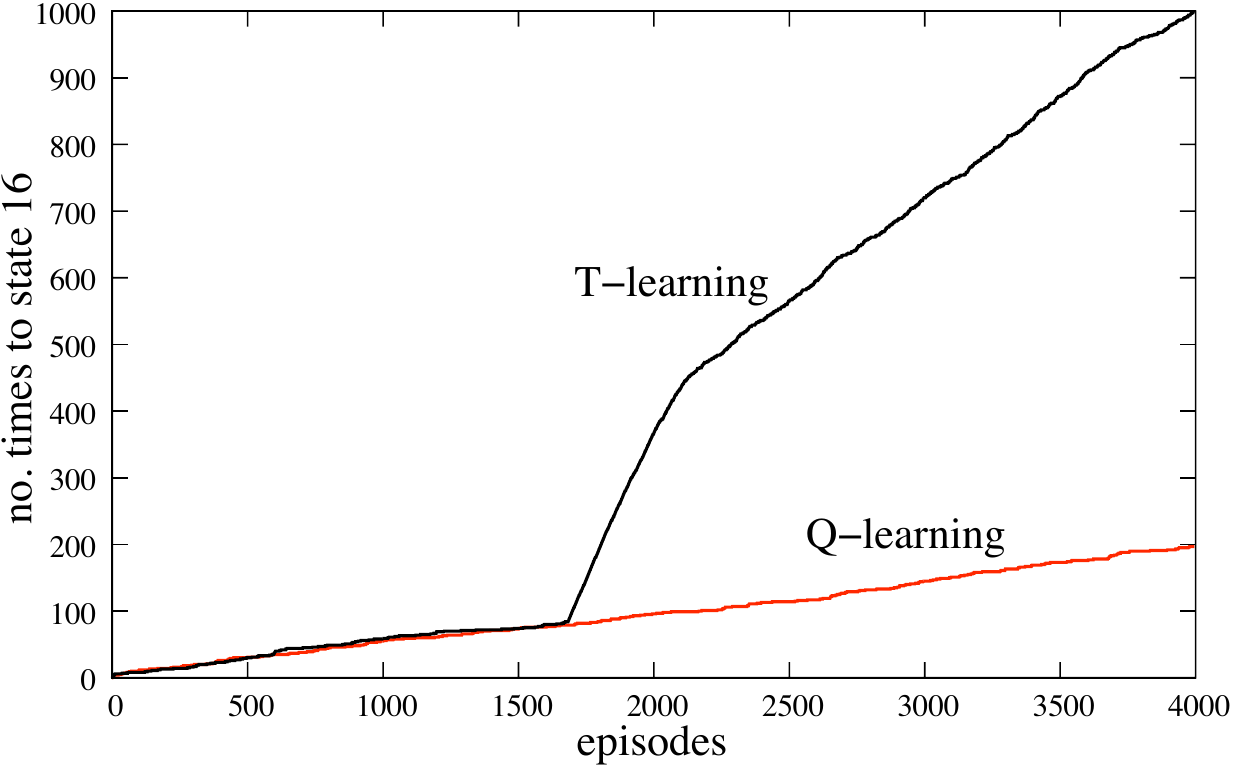}}
\hspace{1cm}
  \subfloat[Convergence of T-values]{\label{fig:tbehave}\includegraphics[width=0.45\textwidth]{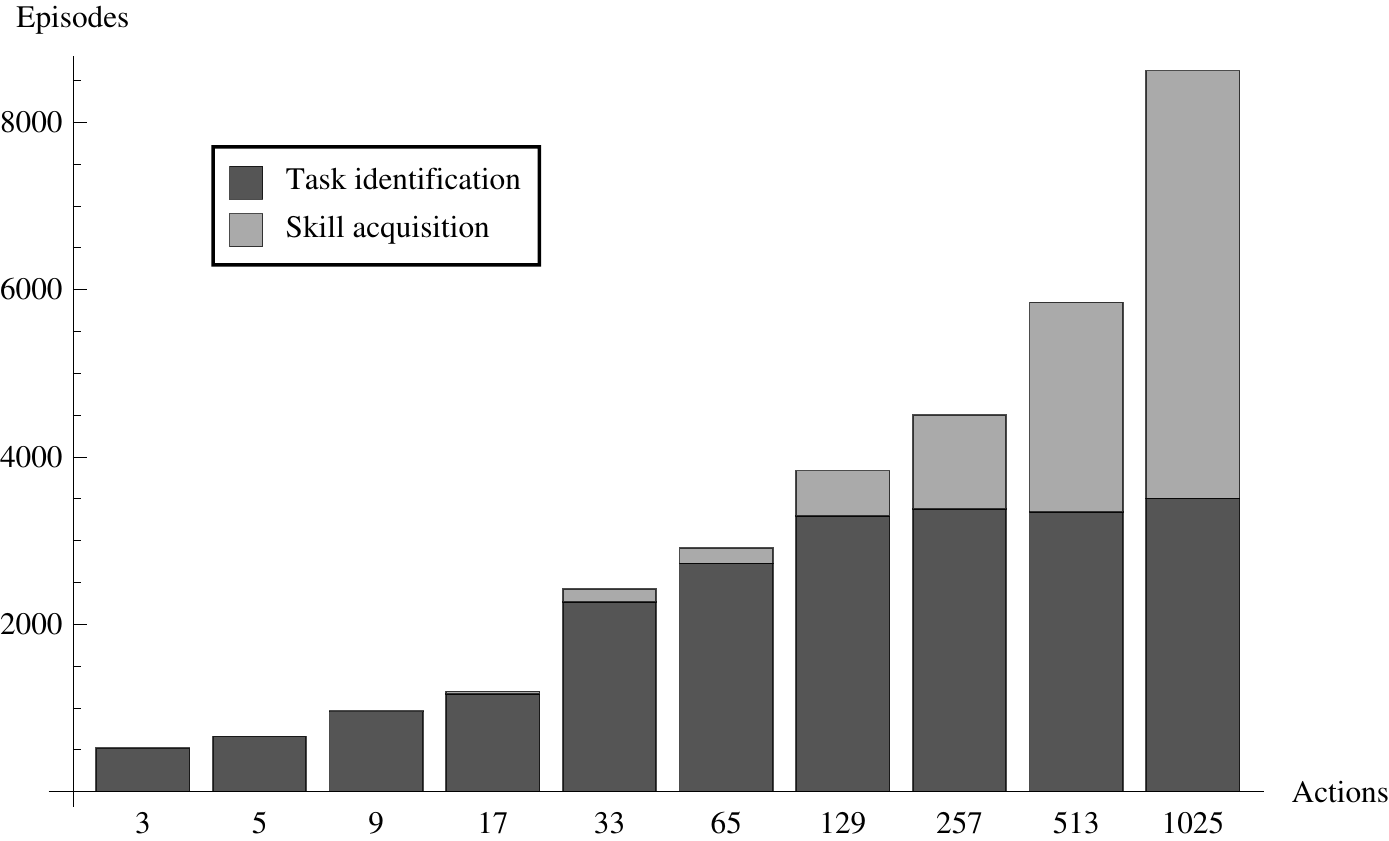}}                
 \caption{Left: Initially both algorithms sample  $1 \to 3$ with the
   same frequency. $T$-learning is fundamentally different than
   $Q$-learning; it can value state transitions before having
   discovered how to reliably make them: In this typical example (for
   $101$ actions) after approximately 1,600 episodes, the $T$-learning
   algorithm prefers $1 \to 3$ over $1 \to 2$. The sharp increase in
   visits to state $16$ occurs after the $T$-values have converged
   and it has yet to learn the skilled actions. After the optimal
   policy is found---skill acquired---further visits to state $16$ are the result of exploration. The slope of the $Q$-learning curve remains constant until convergence to the optimal policy; valuable states and actions must be discovered as pairs. Right: The number of episodes for convergence of $T$-values and convergence of policy are shown for action spaces of varying size. The number of episodes required to identify the task---$T$-values converging---decouples from the size of the action space once the space is sufficiently large. That is, the number of episodes to identify the ideal transitions is independent of the size of the action space!} 
 \label{fig:behave}
\end{figure*}

Even with a large state space, learning functions in $\Ss \times \Ss$ is not as daunting as it seems. Typically the state space is far from fully-connected, so that the sampling needed is
 nowhere near quadratic with respect to the size of the state space. Further, a large action space does not effect the difficultly of learning a function whose domain is $\Ss \times \Ss$. However, when learning functions in the space $\Ss \times \Aa$, sampling must be
 done at all state-action pairs. Learning transition functions has theoretical advantages over learning state value functions: (1) transition functions give more information, values are assigned to the transitions between states, rather than the states themselves, (2) they implicitly contain a model of the environment. As a result of (1) real-world RL function approximation may prove more powerful for transition functions (regardless of the learning rule) than for state functions since there are more relationships to generalize from. %

Robots can be
initialized with certain $T$-values, say learned in simulation, and then be left to learn the
control
 that traverses the valuable states. This is not equivalent to
using properly initialized state-values. Since the $T$-values come with an implicit model
of the environment, the robot, in any given state $s$ has a goal state $s^{*} = \arg\!\max_{\tilde{s}}T(s,\tilde{s})$. The deviation from this goal state after taking an action can be used to learn the relationships between the actions both for the current transition and at other transitions. This is a natural framework for transfer~\cite{transfer}. More generally, 
$T$-learning is biologically plausible in that it
allows a goal-state to be valued highly before possessing the skills
 needed to reach that goal.  After
seeing someone ride a unicycle for the first time it is clear that
this is a skill that can be learned. We can value the difficult goal of
balancing on a unicycle before having ever tried it.

Learning transition values is also very attractive in environments that are non-Markovian. A hidden state may dramatically effect the control
required to achieve specific state transitions without altering the
values of these transitions. For example, a strong wind may
dramatically change the control required for flying a plane without
effecting the desired flight path. Preliminary work has shown that agents
relying on transition values are robust in environments whose dynamics are
non-stationary, in the way suggested above, due to the fact that learning
is invariant with respect to these changes in the transition tables. 

On a more abstract level, rather than focusing on single actions, we can consider
subgoals~\cite{Bakker:04ias} or behaviors~\cite{Konidaris:2007} that
transition the agent between relevant states. As soon as an agent has
discovered a state transition it can assign it a value, regardless of
whether the behavior initially making the transition is reliable.
These values can then be used to drive the agent to interesting or
valuable states where methods from the options framework can be employed
to learn how to reliably reach other valuable states or reward.

\section{Conclusion}

The T-Learning algorithm learns values fundamentally different from
Q-Learning, allowing an agent to quickly identify the valuable transitions in an
environment, regardless of the size of the action space. The behavior
exhibited by T-learning allows an agent to sample from the environment in a way that
amounts to the focused learning of a skill. As a result 
T-Learning, in realistic scenarios, can behave {\em arbitrarily better} than Q-learning.


\bibliographystyle{abbrv}
\bibliography{tlearning}

\end{document}